\documentclass[runningheads]{llncs}

\usepackage{eccv}

\usepackage{eccvabbrv}

\usepackage{graphicx}
\usepackage{booktabs}
\usepackage{bbding}

\usepackage[accsupp]{axessibility}  
\usepackage{multirow}
\usepackage{array}

\usepackage[hidelinks]{hyperref}

\usepackage{orcidlink}

\begin{document}

\title{DiNBV-Grasp: Real-Time, Distance-Aware Two-Stage Next-Best-View Planning for Robotic Grasping} 

\titlerunning{DiNBV-Grasp}

\author{Zilong Xie\inst{1}$^\star$\orcidlink{0009-0001-4436-5322} \and
Jingyu Gong\inst{1,2,3}$^\star$\orcidlink{0000-0002-4536-0953} \and
Xin Tan\inst{1,2}\orcidlink{0000-0001-9346-1196} \and
Zhizhong Zhang\inst{1,3}\textsuperscript{\Envelope}\orcidlink{0000-0001-6905-4478} \and
Yanyun Qu\inst{4}\orcidlink{0000-0002-8926-4162} \and
Lizhuang Ma\inst{1}\orcidlink{0000-0003-1653-4341} \and
Yuan Xie\inst{1,2}\textsuperscript{\Envelope}\orcidlink{0000-0001-6945-7437}}

\authorrunning{Z.~Xie et al.}

\institute{School of Computer Science and Technology, East China Normal University, China \and
Chongqing Key Laboratory of Precision Optics, Chongqing Institute of East China Normal University, China\and
Shanghai Key Laboratory of Computer Software Evaluating and Testing, China\and
School of Informatics, Xiamen University, China\\
\email{zilong\_6037@stu.ecnu.edu.cn} \email{yyqu@xmu.edu.cn}\\
\email{\{jygong, xtan, zzzhang, lzma, yxie\}@cs.ecnu.edu.cn}}

\maketitle

\renewcommand{\thefootnote}{$\star$}
\footnotetext[2]{Equal Contribution \;\; \Envelope\ Corresponding Author.\enspace This version includes minor corrections and additions to the proceedings version.}

\begin{abstract}
Next-Best-View (NBV) planning is widely adopted for robotic grasping in cluttered and occluded environments. However, existing methods typically assume that a single observation at an arbitrary camera–object distance is sufficient for accurate grasp prediction. This overlooks a critical factor: grasp perception quality is highly sensitive to viewing distance, and the optimal distance varies significantly across object categories and scales. To address this limitation, we propose DiNBV-Grasp, a real-time, distance-aware two-stage NBV framework. In Stage I, we distill a large vision-language model (VLM) into a lightweight planner for global semantic NBV exploration, resolving occlusions and generating an initial grasp hypothesis with low-latency inference. In Stage II, we introduce a local distance-aware refinement mechanism that explicitly models the relationship between viewing distance and grasp success probability. During manipulator approach, a lightweight scoring network evaluates intermediate observations and selects the most informative one for final grasp prediction. Extensive simulation and real-world experiments demonstrate that DiNBV-Grasp consistently improves grasp success rates across diverse objects, particularly for small and occluded targets, while preserving real-time performance, highlighting the importance of distance-aware perception in NBV-based grasping. Our code is available at \url{https://github.com/XieZilongAI/DiNBV-Grasp}.
\end{abstract}

\begin{figure*}[t]
    \centering
    \includegraphics[width=\textwidth]{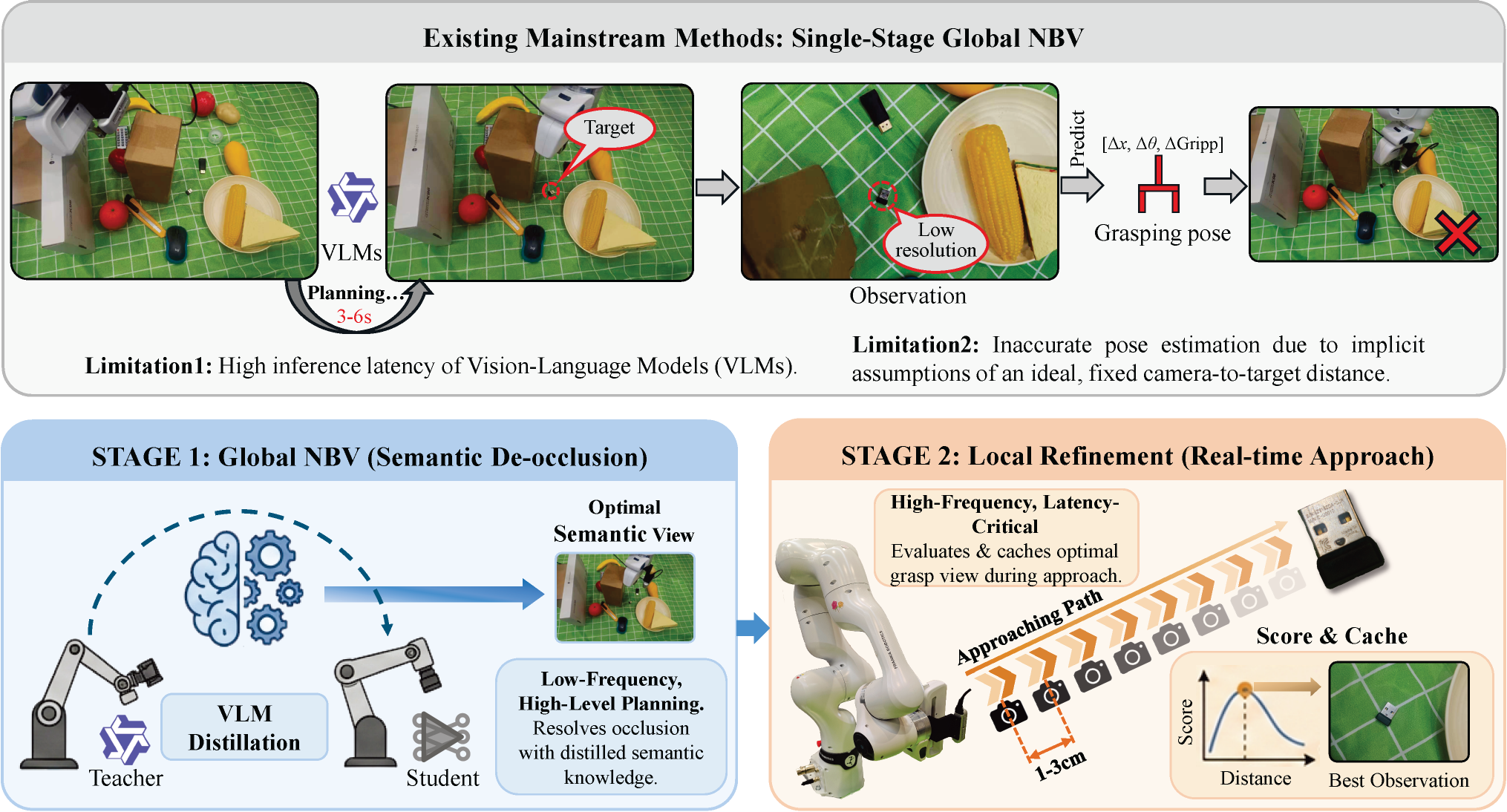}
    \caption{\textbf{Motivation of DiNBV-Grasp.} 
Top: existing single-stage NBV pipelines rely on a single globally selected viewpoint, suffering from high VLM latency and distance-agnostic grasp failures. 
Bottom: the proposed two-stage framework decouples global semantic de-occlusion from local distance-aware refinement, enabling real-time and robust grasping under varying observation distances.}
    \label{fig:introduction}
\end{figure*}

\section{Introduction}

Reliable robotic grasping in cluttered and unstructured scenes remains challenging due to occlusions and incomplete observations~\cite{GraspNet1Billion2020,lauri2020multi,breyer2022closed}. Next-Best-View (NBV) planning actively moves the sensor to improve visibility and grasp prediction quality. However, most existing NBV approaches~\cite{jin2023neu,wang2025graspview,shi2025viso} simplify the pipeline to selecting a single globally optimal viewpoint followed by grasp prediction from that observation. Similarly, object-centric sparse prediction~\cite{li2024manipllm,li2025object} and keyframe-based methods~\cite{eisner2022flowbot3d,zhong20233d} implicitly assume the existence of a suitable camera–object distance. In practice, grasp prediction quality is highly sensitive to viewing distance, especially for small or geometrically complex objects.

Another limitation lies in computational latency. Recent works~\cite{zhang2023affordance,wilson2025pop,strong2025next,liu2026activevla} incorporate 3D reconstruction~\cite{kerbl20233d,must3rcvpr25,eccvLatentSplat} or large vision-language models (VLMs)~\cite{radford2021learning,jia2021scaling,li2023blip,liu2023visual,alayrac2022flamingo} to enhance semantic reasoning. While effective, these approaches~\cite{wang2025graspview,shi2025viso} typically incur second-level inference delays, making it difficult to support continuous robotic control, as illustrated in Fig.~\ref{fig:introduction} (top). Forcing a single heavy model to handle both global semantic reasoning and high-frequency local decision-making is therefore impractical.

To address these issues, we propose DiNBV-Grasp, a two-stage NBV framework that explicitly models distance-aware grasp perception (Fig.~\ref{fig:introduction}, bottom). The first stage performs semantic-aware global exploration. We distill a large VLM offline into a lightweight student that scores candidate views in 28 ms per image in our ablation. 

The second stage operates during manipulator approach. A lightweight distance-aware scorer evaluates intermediate observations using RGB features, depth statistics, viewing distance, and object category information, and selects the most informative observation for final grasp prediction. This decoupling allows global occlusion resolution and local high-frequency refinement to operate under distinct computational constraints.

In simulation, Stage II improves ASR by 2.41--5.89 percentage points over the heuristic selector across five manipulation methods and both test splits. In five cluttered real-world scenes, the full system achieves or ties for the highest success rate among the evaluated methods, averaging 2.90 Stage I observations per trial.

Our main contributions are threefold:
(1) We identify and systematically analyze the critical influence of camera–object distance on grasp perception quality, highlighting a previously underexplored dimension in NBV-based manipulation. 
(2) We propose DiNBV-Grasp, a two-stage framework with lightweight view scoring that decouples global semantic exploration from local distance-aware refinement, enabling adaptive selection of informative observations during manipulator approach. 
(3) We evaluate Stage II as a plug-in in simulation and the full two-stage system in five cluttered real-world scenes.

\section{Related Work}

\noindent\textbf{Object-Centric Grasping.} 
Object-centric grasping methods predict grasp poses directly from visual observations~\cite{eisner2022flowbot3d,zhong20233d,li2024manipllm,li2025object,wei2025mathcal}. End-to-end 6-DOF networks such as Contact-GraspNet~\cite{Contactgraspnet2021} and VGN~\cite{Breyer2021VolumetricGN} generate grasp candidates from RGB-D inputs or voxelized representations in real time. Related 3D pose and shape estimation methods also exploit monocular cues or geometric features to recover object or hand structure under occlusion~\cite{zhang20243d,liu20246dof}, but they do not address active view selection during grasp execution. These approaches typically assume that sufficiently complete object geometry is observable from a single viewpoint under a suitable camera distance. As a result, their performance degrades under heavy occlusion or extreme near-field observations where geometric evidence becomes incomplete or distorted.

Recent efforts aim to improve robustness through structured or semantic representations. STORM~\cite{chapin2026storm} and OmniManip~\cite{pan2025omnimanip} learn object abstractions from primitives or semantic slots to enhance generalization. Interaction-centric frameworks such as FOCUS~\cite{ferraro2025focus} and AnyDexGrasp~\cite{fang2025anydexgrasp} further emphasize identifying task-relevant entities in cluttered environments, while SimToReal~\cite{lin2025sim} extends grasping to tool-oriented manipulation scenarios. Point-cloud completion methods such as FACNet~\cite{yu2025facnet} improve shape recovery from partial observations, but still rely on the available input view rather than optimizing camera motion. Although these methods improve perception and reasoning under clutter, they still rely on pre-acquired observations and generally lack adaptive viewpoint or distance modeling during execution. Keyframe-based strategies partially mitigate occlusion but do not explicitly optimize observation distance throughout the grasping process.

\noindent\textbf{Active Perception for Grasping.} 
Active perception addresses visual uncertainty by planning informative viewpoints~\cite{breyer2022closed,ma2024active,vutetakis2025active}. Typical NBV strategies optimize information gain, entropy reduction, or grasp prediction uncertainty~\cite{shi2025viso}. For example, GraspView~\cite{wang2025graspview} employs VLM-based scoring to evaluate rendered candidate views, while Neural Graspness Field~\cite{ma2024active} predicts spatial grasp distributions to guide camera motion via grasp inconsistency. Other systems, such as ActiveUMI~\cite{zeng2025activeumi} and RASP~\cite{liu2025rasp}, incorporate joint viewpoint planning and depth completion inspired by human exploration strategies. More recent cognitive frameworks like CoMe-VLA~\cite{li2026act} formulate active perception as a non-Markovian decision process for long-horizon tasks.

Despite these advances, most NBV-based methods focus primarily on global exploration to resolve occlusion before executing a final grasp from a single selected observation. The camera–object distance during the approach phase is typically treated as incidental rather than an explicit optimization variable, leaving its influence on grasp prediction quality insufficiently modeled.

\noindent\textbf{Foundation Models in Robotic Perception and Planning.} 
Foundation Models, particularly Vision–Language Models (VLMs), have been increasingly integrated into robotic systems for open-vocabulary perception, semantic reasoning, and task planning~\cite{radosavovic2023real,liu2023partslip,xu2024rt}. VISO-Grasp~\cite{shi2025viso} leverages VLMs to reason about spatial relationships between targets and occluders for semantic-driven NBV planning. Halo~\cite{shou2026halo} and InstructVLA~\cite{yang2025instructvla} introduce embodied reasoning chains to align perception with downstream actions.

Related foundation-model advances also include adaptive prompting of CLIP~\cite{radford2021learning} for dense scene parsing~\cite{li2024clip}, depth foundation models for generalizable visual depth estimation~\cite{xu2026towards}, lifelong skill scaling~\cite{zhang2025t2s}, action-aligned affordance planning~\cite{wang2026afford}, and semantic-geometric mapping for embodied navigation~\cite{li2026bridging}.

However, large-scale VLM inference incurs substantial computational overhead~\cite{wang2025graspview}, limiting real-time applicability in closed-loop control. Consequently, recent research has explored distillation and model compression. BitVLA~\cite{wang2025bitvla} proposes a native 1-bit architecture for edge deployment, while Drive-KD~\cite{lian2026drive} and Shallow-$\pi$~\cite{jeon2026shallow} adopt multi-teacher distillation and architectural pruning to improve inference efficiency. Nevertheless, most FM-based grasping systems operate primarily on 2D image inputs without explicitly incorporating 3D spatial priors or depth-aware modeling~\cite{sripada2024ap}. 

These observations suggest that while foundation models enhance semantic reasoning, their integration into real-time robotic grasping requires lightweight designs and tighter coupling with geometry-aware active perception mechanisms.

\section{Method}

\subsection{Problem Formulation}

Let $\mathcal{V}=\{v_k\}_{k=1}^{N}$ denote a discrete set of candidate camera viewpoints in the robot workspace, where each $v_k \in SE(3)$ represents a 6-DoF camera pose. 
The robot incrementally observes the environment and constructs a global scene representation $\mathcal{S}$.

Given a target object specified by a textual query $T$, the goal is to actively select informative viewpoints and execute a successful grasp.

Our system operates in two stages: (1) global NBV planning, which iteratively selects an informative viewpoint $v_{\text{global}}\in\mathcal{V}$ and predicts an initial grasp hypothesis $\pi_0$, and (2) distance-aware local view selection, which evaluates observations collected during manipulator approach and selects the most informative one to refine the final grasp $\pi^\star$.

\begin{figure*}[t]
    \centering
    \includegraphics[width=\textwidth]{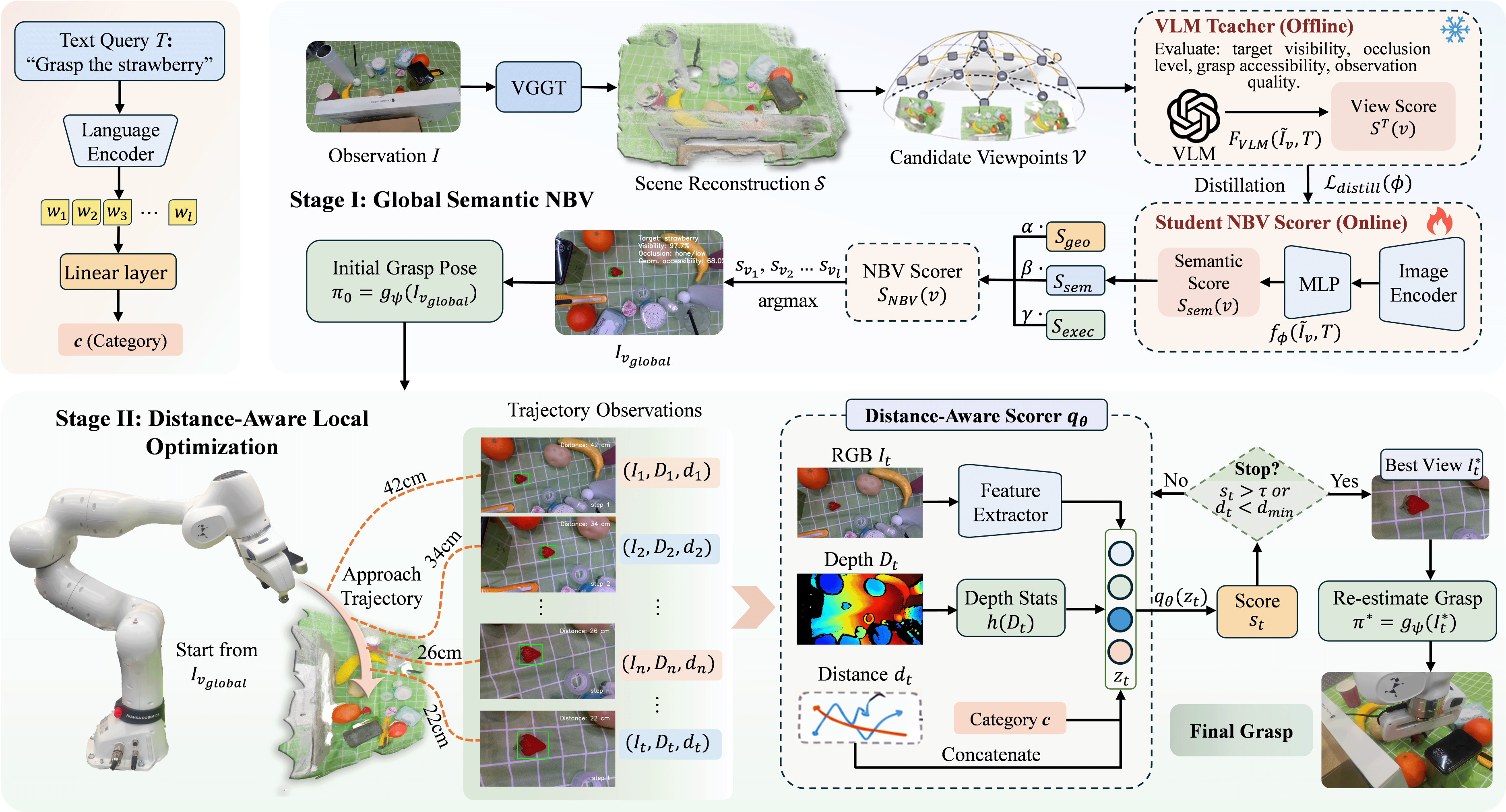}
    \caption{\textbf{Framework of DiNBV-Grasp.} 
Stage I performs global semantic NBV planning via VLM distillation to resolve occlusions and generate an initial grasp hypothesis. 
Stage II conducts high-frequency distance-aware view scoring during manipulator approach and selects the most informative observation for final grasp prediction.}
    \label{fig:framework}
\end{figure*}

\subsection{Scene Reconstruction and Multi-View Rendering}

Following~\cite{wang2025graspview}, the system uses VGGT~\cite{wang2025vggt} to reconstruct a global scene model $\mathcal{S}$ from accumulated RGB-D observations. The resulting scene representation supports rendering from candidate camera poses.

\noindent\textbf{Candidate viewpoints.}
Candidate camera poses are sampled on a hemispherical shell around a fixed tabletop search region in the robot workspace, with discretized radii and viewing angles. The candidate set $\mathcal{V}$ is available before the target is detected. Each candidate pose is validated via inverse kinematics and collision checking.

\noindent\textbf{View rendering and evaluation.}
For each $v\in\mathcal{V}$, a synthetic observation is rendered:

\begin{equation}
\tilde{I}_v = \mathcal{R}(\mathcal{S}, v).
\end{equation}

These virtual observations are evaluated without moving the camera, and the robot executes motion only to the selected next-best-view.

\subsection{Global NBV via VLM Distillation}

At exploration step $t$, the next-best-view is selected by maximizing a visibility-aware utility function that evaluates geometric visibility, semantic relevance, and execution feasibility:

\begin{equation}
S_{\text{NBV}}(v,t)
=
\alpha_t S_{\text{geo}}(v)
+
\beta_t S_{\text{sem}}(v)
+
\gamma_t S_{\text{exec}}(v),
\end{equation}

\noindent where $(\alpha_t,\beta_t,\gamma_t)$ are visibility-adaptive weights conditioned on the current target visibility $V_t$.
All three score terms are normalized over the feasible candidate set at step $t$:
$S_{\text{geo}}$ measures how much previously unobserved space a candidate view can probe,
$S_{\text{sem}}$ is the distilled teacher-VLM score measuring target visibility, semantic relevance, and graspability,
and $S_{\text{exec}}$ encodes rule-based execution feasibility, including inverse-kinematics validity, collision checking, and safety constraints.
When the target is heavily occluded, the utility emphasizes geometric exploration; as visibility improves, semantic and execution terms receive higher influence for selecting grasp-relevant and physically reachable views.
The selected viewpoint is:

\begin{equation}
v_{\text{global}}
=
\arg\max_{v\in\mathcal{V}} S_{\text{NBV}}(v,t).
\end{equation}

Directly computing semantic utility with large vision--language models is computationally expensive. 
We therefore distill the scoring behavior of a teacher VLM into a lightweight student predictor.
A teacher VLM produces semantic scores:

\begin{equation}
S^{T}(v)
=
F_{\text{VLM}}(\tilde I_v, T),
\end{equation}

\noindent and a compact model $f_\phi$ is trained to approximate this mapping:

\begin{equation}
\mathcal{L}_{\text{distill}}(\phi)
=
\mathbb{E}_{v\sim\mathcal{V}}
\left[
\ell\big(
S^{T}(v),
f_\phi(\tilde I_v, T)
\big)
\right].
\end{equation}

Until the target is first detected in an acquired RGB-D observation, $S_{\text{sem}}(v)=0$ for all candidates. Thereafter, the student scores the rendered views:

\begin{equation}
S_{\text{sem}}(v)
=
f_\phi(\tilde I_v, T).
\end{equation}

Global NBV planning is executed in a closed-loop manner. At each step, the system renders candidate views from the current scene representation, evaluates their utilities, and selects the next-best-view $v_{\text{global}}$. The robot moves to the selected viewpoint, captures a new observation $I_{v_{\text{global}}}$, and updates the scene representation $\mathcal{S}$.

This process terminates deterministically when either exploration reaches $N_{\max}$ steps after $I_0$ or the target visibility $V_t$ reaches a threshold $\eta_v$.

After exploration terminates, the system selects the final global observation and predicts an initial grasp hypothesis:

\begin{equation}
\pi_0 = g_\psi(I_{v_{\text{global}}}),
\end{equation}

\noindent where $g_\psi$ is a grasp prediction network. Stage II begins only when a valid global observation yields an initial grasp hypothesis $\pi_0$ that supports a feasible approach. Otherwise, the trial is safely aborted.

\subsection{Distance-Aware Local View Selection During Approach}

After the global exploration stage, the manipulator approaches the target along a collision-free trajectory. 
During this motion, the system continuously evaluates intermediate observations and selects the most informative one for final grasp prediction.

\noindent\textbf{Observation representation.} 
At each approach step $t$, the robot acquires an RGB-D observation and measures the current camera--object distance $d_t$. 
Each observation is encoded into a compact feature vector:

\begin{equation}
z_t
=
\big[
\phi_{\text{rgb}}(I_t),\;
h(D_t),\;
\hat d_t,\;
c
\big],
\end{equation}

\noindent where $\phi_{\text{rgb}}$ is a visual feature extractor, $h(D_t)$ denotes low-dimensional depth statistics, $\hat d_t$ is the normalized distance, and category encoding $c$ is obtained using a pre-trained CLIP text encoder, which converts the category name into a 512-dimensional text embedding. Only the text encoder is used in our implementation, enabling the representation to generalize to unseen categories without relying on a fixed vocabulary. We compute $h(D_t)$ from numeric depth measurements. Colorized depth maps are used only for visualization.

\noindent\textbf{Distance-aware success scoring.}
We model view quality using a lightweight distance discriminator:

\begin{equation}
p_t
=
q_\theta(z_t) \in [0,1],
\end{equation}

\noindent which predicts the probability that a grasp generated from the current observation will succeed.
Unlike heuristic distance selection, this model directly estimates action success likelihood conditioned on visual evidence, object category, and viewing distance.

\noindent\textbf{Online selection.}
During the approach, the system maintains the best observation encountered so far:

\begin{equation}
t^\star_T
=
\arg\max_{1 \le t \le T} p_t.
\end{equation}

Observation sampling stops when the highest score observed so far reaches $\tau$ or the camera reaches the minimum observation distance $d_{\min}$:

\begin{equation}
\text{stop if }
\big(
\max_{1\le t\le T} p_t \ge \tau
\big)
\;\;\text{or}\;\;
d_T \le d_{\min}.
\end{equation}

The final grasp pose is estimated from the highest-scoring observation acquired by the stopping step:

\begin{equation}
\pi^\star
=
g_\psi(I_{t^\star_T}).
\end{equation}

This mechanism enables continuous view optimization during motion and explicitly accounts for the dependence of grasp success on viewing distance.

\subsection{Training Objective}

The global semantic student $f_\phi$ is trained using the distillation loss defined previously. 
Here we describe the learning objective for the distance-aware discriminator $q_\theta$.

\noindent\textbf{Supervision.}
Each training sample consists of an observation and the outcome of an independently executed grasp.
For each scene, we reset the robot and scene to the same initial state at each sampled viewing distance, predict a grasp from the current observation alone, and keep the grasp predictor and control settings fixed.
The outcome of each execution provides the binary label $y_t \in \{0,1\}$ for the corresponding observation feature $z_t$, and the discriminator is trained to predict grasp success conditioned on the observation.

\noindent\textbf{Sample-level success prediction.}
We first optimize a binary classification objective using cross-entropy loss:

\begin{equation}
\mathcal{L}_{\text{bce}}
=
\mathbb{E}_t
\Big[
- y_t \log q_\theta(z_t)
-
(1-y_t)\log(1-q_\theta(z_t))
\Big].
\end{equation}

This term supervises the observation-level score using binary execution outcomes.

\noindent\textbf{Scene-level distance ranking.}
For observations collected from the same scene, successful viewing distances should receive higher scores than unsuccessful ones.
We therefore impose a ranking constraint:

\begin{equation}
\mathcal{L}_{\text{rank}}
=
\max
\big(
0,\;
m - (\bar p_{\text{succ}} - \bar p_{\text{fail}})
\big),
\end{equation}

\noindent where $\bar p_{\text{succ}}$ and $\bar p_{\text{fail}}$ denote the mean predicted success probabilities over successful and failed distances within the same scene, and $m$ is a margin.

\noindent\textbf{Pairwise contrastive separation.}
To further enforce discriminability between good and poor viewing distances, we apply a pairwise margin ranking loss on sampled success--failure pairs from the same scene:

\begin{equation}
\mathcal{L}_{\text{ctr}}
=
\mathbb{E}_{(i,j)}
\max\big(0,\; m - (p_i - p_j)\big),
\end{equation}

\noindent where $i$ and $j$ index successful and failed observations respectively.

\noindent\textbf{Overall objective.}
The final training objective for the local scorer is:

\begin{equation}
\mathcal{L}_{\text{scorer}}(\theta)
=
\mathcal{L}_{\text{bce}}
+
\lambda_1 \mathcal{L}_{\text{rank}}
+
\lambda_2 \mathcal{L}_{\text{ctr}},
\end{equation}

\noindent where $\lambda_1$ and $\lambda_2$ balance scene-level ordering and pairwise discrimination.

The global student is trained with $\mathcal{L}_{\text{distill}}$, and the local scorer is trained with $\mathcal{L}_{\text{scorer}}$.


\section{Experiments}

We evaluate the proposed method in both large-scale simulations and real-world robotic experiments to validate its algorithmic effectiveness and practical robustness.

\subsection{Simulation Setup}

Experiments are conducted in interactive environments built on the SAPIEN physics simulator with PartNet-Mobility assets~\cite{xiang2020sapien}. Manipulation is performed using a Franka Panda arm with a suction gripper. Our simulation data span 20 object categories. The Stage II dataset contains approximately 40,000 successful and 40,000 failed independent grasp attempts, with a 1:1 success-to-failure ratio in the training set. The 20 categories are divided into 12/4/4 for training, validation, and testing, respectively, with the test categories used for \textit{Test Unseen}. Within the training categories, object instances are split into 70\%/15\%/15\% for training, validation, and testing, respectively, with the test instances used for \textit{Test Seen}. Data splitting is performed at the object-instance, scene, and trajectory levels: each object instance, together with its associated scenes and trajectories, is assigned to only one split.

\subsection{Real-World Experimental Setup}

The physical platform consists of a 7-DoF Franka Emika Panda manipulator and an Intel RealSense D435i RGB-D camera. The system is implemented in Python/PyTorch, utilizing ROS~2 for middleware and MoveIt2 for motion planning. Grasp execution is powered by Contact-GraspNet~\cite{Contactgraspnet2021}. For real-world Stage II, we fine-tune the simulation-trained local scorer on a small set of real-world images that we collected. Each object is evaluated over 10 independent trials.

\subsection{Evaluation Metrics} 

To evaluate our framework comprehensively, we employ a consistent set of metrics across both simulation and real-world experiments. We report Average Success Rate (ASR, \%) for simulation experiments and Success Rate (SR, \%) for real-world experiments. In simulation, ASR is defined as the percentage of successful grasps where the target displacement exceeds $\delta = 0.1$ after execution. In real-world experiments, SR and Failure Rate (FR) are the fractions of the 10 trials ending in grasp success and grasp failure, respectively.
Abort Rate (AR, \%) quantifies the percentage of trials where the system fails to identify a valid observation within the maximum number of exploration steps $N_{\max}$. Average Views (\#Views) is the mean number of Stage I observations per trial, counting $I_0$ toward the $N_{\max}+1=6$ limit. Regarding temporal performance, we report the Average Frame Time (ms) in simulation to measure inference latency per frame, and the Average Total Time (s) in real-world experiments to measure the overall time per trial. Timing starts upon receiving the initial RGB-D observation and ends upon completion of Stage II; only trials reaching Stage II are included in the average.

\subsection{Implementation Details}

Stage I runs for up to $N_{\max}=5$ steps after $I_0$ and stops early when $V_t$ meets the visibility threshold of 0.75. Entry into Stage II also requires a feasible initial grasp hypothesis. The local distance-aware scorer uses $\tau=0.8$ as its score threshold during approach. The distance-aware discriminator takes a 1035-dimensional input comprising RGB features (512D), depth statistics (10D), normalized distance (1D), and a category encoding (512D). In simulation, a linear projection maps the 768-dimensional output of DINOv3-Base~\cite{simeoni2025dinov3} to 512 dimensions. We set $d_{\min}=15$\,cm as the minimum observation distance during approach in our setup. The training objective combines a scene-level ranking loss ($\lambda_1=1.0$) and a pairwise contrastive loss ($\lambda_2=0.5$, margin $m=0.2$). Optimization is performed using Adam with learning rate $10^{-4}$, batch size 128, for 50 epochs. The RGB encoder remains frozen during training, and distances are normalized to $[0,1]$ within the sampled range. For VLM distillation, Qwen3-VL-32B-Instruct~\cite{bai2025qwen3} is used as the teacher model to generate semantic supervision for candidate rendered views. The student concatenates DINOv3 image features with an encoding of $T$ and feeds the combined feature to a small MLP scoring head.

\subsection{Baselines and Integration Protocol}

In simulation, we compare against Flowbot3D~\cite{eisner2022flowbot3d}, Implicit3D~\cite{zhong20233d}, ManipLLM~\cite{li2024manipllm}, CrayonRobo~\cite{li2025object}, and RobMRAG~\cite{robmrag}, which span geometric, learning-based, and vision-language paradigms. All assume manipulation prediction from a single observation at a suitable distance. During execution, observations acquired at different distances are evaluated and the one with highest predicted manipulation relevance is selected for final pose estimation. Additional strategies include \emph{random} (random observation selection), \emph{nearest} (closest-distance selection), and \emph{heuristic}. The heuristic baseline selects the observation distance using a weighted combination of a Gaussian prior centered at a category-dependent preferred distance and a depth-quality score measuring depth consistency and valid sensing range.

In real-world experiments, we compare against \emph{initial-view}, which predicts grasp from a predefined initial camera pose only; \emph{fixed-traj.}, where the robot follows fixed circular trajectories centered on the target object at a $30^\circ$ depression angle and uniformly samples up to $N_{\max}$ additional views, terminating early upon confident detection; and the method of \textit{Breyer} et al.~\cite{breyer2022closed}, a closed-loop next-best-view strategy based on geometric information gain for grasp planning. All methods share identical hardware configuration, planning stack, and execution interface to ensure fair comparison. 

\begin{table}[tb]
\centering
\small
\setlength{\tabcolsep}{5pt}
\caption{Simulation results (ASR, \%) with our distance-aware second-stage plug-in on object-centric grasping methods. \emph{Published} values are taken from the cited papers. Random, Nearest, Heuristic, and Ours use the same multi-distance evaluation protocol.}
\label{tab:simulation_results}
\begin{tabular}{l|ccccc}
\toprule
\multicolumn{6}{c}{\textbf{Test Seen Split}} \\
\midrule
 & Random & Nearest & Published & Heuristic & \textbf{Ours} \\
\midrule
Flowbot3D~\cite{eisner2022flowbot3d}   & 38.88 & 29.76 & 37.00 & 40.09 & \textbf{45.16} \\
Implicit3D~\cite{zhong20233d}  & 46.45 & 36.66 & 46.00 & 47.58 & \textbf{53.47} \\
ManipLLM~\cite{li2024manipllm}   & 54.90 & 41.37 & 56.00 & 61.58 & \textbf{65.62} \\
CrayonRobo~\cite{li2025object}  & 72.52 & 59.94 & 74.00 & 79.92 & \textbf{83.01} \\
RobMRAG (\textit{zero-shot})~\cite{robmrag}    & 45.24 & 43.16 & 49.25 & 52.61 & \textbf{57.69} \\
\midrule
\multicolumn{6}{c}{\textbf{Test Unseen Split}} \\
\midrule
 & Random & Nearest & Published & Heuristic & \textbf{Ours} \\
\midrule
Flowbot3D~\cite{eisner2022flowbot3d}   & 31.92 & 25.23 & 32.00 & 34.03 & \textbf{38.53} \\
Implicit3D~\cite{zhong20233d}  & 39.00 & 30.65 & 41.00 & 44.32 & \textbf{46.73} \\
ManipLLM~\cite{li2024manipllm}   & 50.25 & 31.24 & 51.00 & 55.60 & \textbf{58.42} \\
CrayonRobo~\cite{li2025object}  & 70.56 & 55.44 & 72.00 & 75.88 & \textbf{79.67} \\
RobMRAG (\textit{zero-shot})~\cite{robmrag}   & 44.01 & 34.52 & 43.53 & 46.24 & \textbf{51.76} \\
\bottomrule
\end{tabular}
\end{table}

\subsection{Simulation Results}

Table~\ref{tab:simulation_results} reports results on the Test Seen and Test Unseen splits. We average our runs over three random seeds. Under the shared multi-distance protocol, our selector achieves higher ASR than Random, Nearest, and Heuristic for all five manipulation methods on both splits. The improvement over Heuristic ranges from 2.41 to 5.89 percentage points.

\begin{figure*}[t]
    \centering
    \includegraphics[width=\textwidth]{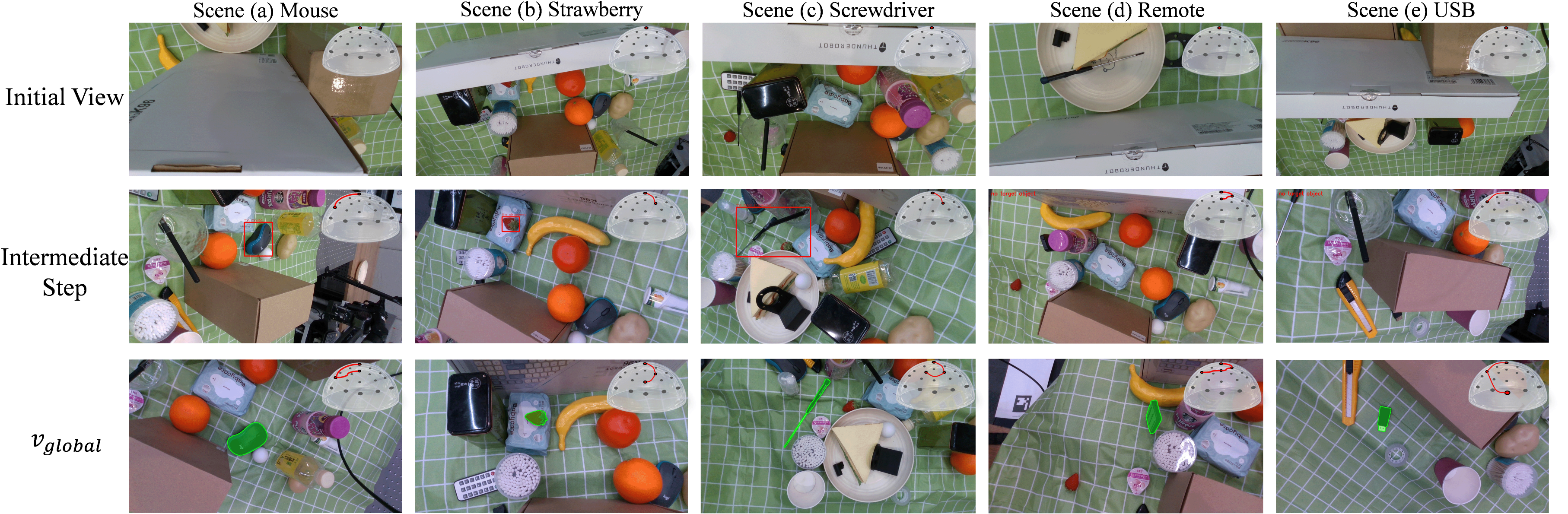}
    \caption{\textbf{Stage I: Global View Exploration.} 
Columns correspond to five cluttered real-world scenes (a–e) in Table~\ref{tab:real_world_results}. 
Rows show RGB-D observations acquired at the initial pose (top), during exploration (middle), and at the selected global pose $v_{\text{global}}$ (bottom). The final observation is used to generate the initial grasp hypothesis.}
    \label{fig:stage1_vis}
\end{figure*}

\begin{table*}[tb]
\centering
\caption{Real-world experimental results in highly cluttered and occluded scenes.}
\label{tab:real_world_results}
\begin{tabular}{
c
>{\centering\arraybackslash}p{2.6cm}
p{3.0cm}
c@{\hspace{6pt}}c@{\hspace{6pt}}c
c
c}
\toprule
Scene & Instruction & Method & SR & FR & AR & \#Views & Time/s \\
\midrule

\multirow{5}{*}{a}
& \multirow{5}{=}{\centering Grasp the mouse}
& \emph{initial-view} & 0/10 & 10/10 & 0/10 & 1.00 & 8.25 \\
& & \emph{fixed-traj.} & 2/10 & 3/10 & 5/10 & 4.50 & 13.52 \\
& & \emph{Breyer et al.} & 4/10 & 3/10 & 3/10 & 4.10 & 24.81 \\
\cmidrule(lr){3-8}
& & \emph{Ours (w/o Stage II)} & 6/10 & 4/10 & 0/10 & 2.20 & 10.82 \\
& & \emph{Ours (Full)} & \textbf{6/10} & 4/10 & 0/10 & \textbf{2.20} & 12.45 \\
\midrule

\multirow{5}{*}{b}
& \multirow{5}{=}{\centering Take the strawberry}
& \emph{initial-view} &  1/10 & 9/10 & 0/10 & 1.00 & 6.82 \\
& & \emph{fixed-traj.} & 1/10 & 3/10 & 6/10 & 4.80 & 11.45 \\
& & \emph{Breyer et al.} & 3/10 & 2/10 & 5/10 & 4.50 & 28.34 \\
\cmidrule(lr){3-8}
& & \emph{Ours (w/o Stage II)} & 4/10 & 4/10 & 2/10 & 3.00 & 12.15 \\
& & \emph{Ours (Full)} & \textbf{5/10} & 3/10 & 2/10 & \textbf{3.00} & 13.72 \\
\midrule

\multirow{5}{*}{c}
& \multirow{5}{=}{\centering Catch the screwdriver}
& \emph{initial-view} &  0/10 & 10/10 & 0/10 & 1.00 & 8.05 \\
& & \emph{fixed-traj.} & 0/10 & 3/10 & 7/10 & 5.00 & 12.10 \\
& & \emph{Breyer et al.} & 1/10 & 5/10 & 4/10 & 4.20 & 26.52 \\
\cmidrule(lr){3-8}
& & \emph{Ours (w/o Stage II)} & 2/10 & 6/10 & 2/10 & 2.60 & 11.42 \\
& & \emph{Ours (Full)} & \textbf{4/10} & 4/10 & 2/10 & \textbf{2.60} & 12.98 \\
\midrule

\multirow{5}{*}{d}
& \multirow{5}{=}{\centering Pick up the remote controller}
& \emph{initial-view} &  0/10 & 10/10 & 0/10 & 1.00 & 7.55 \\
& & \emph{fixed-traj.} & 3/10 & 2/10 & 5/10 & 4.50 & 13.25 \\
& & \emph{Breyer et al.} & 3/10 & 3/10 & 4/10 & 3.80 & 22.44 \\
\cmidrule(lr){3-8}
& & \emph{Ours (w/o Stage II)} & 5/10 & 5/10 & 0/10 & 2.30 & 9.95 \\
& & \emph{Ours (Full)} & \textbf{6/10} & 4/10 & 0/10 & \textbf{2.30} & 11.40 \\
\midrule

\multirow{5}{*}{e}
& \multirow{5}{=}{\centering Retrieve the USB drive}
& \emph{initial-view} &  0/10 & 10/10 & 0/10 & 1.00 & 6.52 \\
& & \emph{fixed-traj.} & 0/10 & 2/10 & 8/10 & 5.60 & 11.88 \\
& & \emph{Breyer et al.} & 0/10 & 4/10 & 6/10 & 4.90 & 30.12 \\
\cmidrule(lr){3-8}
& & \emph{Ours (w/o Stage II)} & 3/10 & 3/10 & 4/10 & 4.40 & 13.82 \\
& & \emph{Ours (Full)} & \textbf{5/10} & 1/10 & 4/10 & \textbf{4.40} & 15.28 \\
\bottomrule
\end{tabular}
\end{table*}

\subsection{Real-world Results}

\begin{figure*}[t]
    \centering
    \includegraphics[width=\textwidth]{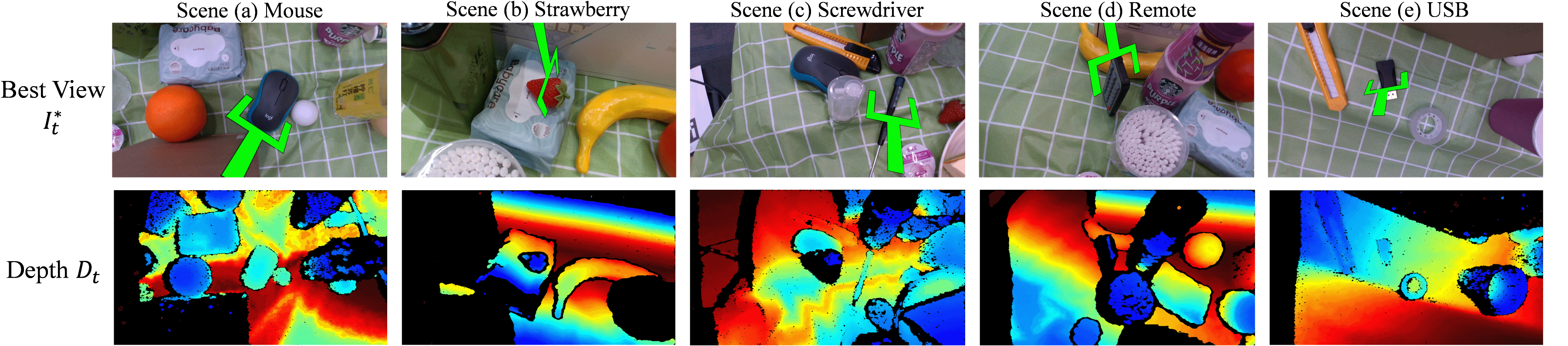}
    \caption{\textbf{Stage II: Distance-Aware Local Refinement.} 
Starting from the global view selected in Stage I, the robot evaluates intermediate observations during approach. 
The top and bottom rows show the selected RGB observation and its corresponding depth map used for final grasp prediction.}
    \label{fig:stage2_vis}
\end{figure*}

We evaluate the proposed framework in five highly cluttered real-world scenarios (Scenes a–e). Figure~\ref{fig:stage1_vis} shows how the target becomes visible as the camera moves. Given a feasible initial grasp hypothesis, Stage II selects the highest-scoring RGB-D observation acquired during approach for final grasp prediction (Fig.~\ref{fig:stage2_vis}).

Quantitative results in Table~\ref{tab:real_world_results} show that \emph{Ours (Full)} achieves or ties for the highest Success Rate (SR) in all five scenes, with an overall SR of 52\% (26/50). Compared to geometric NBV~\cite{breyer2022closed}, our method improves SR in every scene while reducing per-scene \#Views from 3.80--4.90 to 2.20--4.40. Our method averages only 2.90 views across scenes. Average total task times are also lower in every scene (11.40--15.28\,s for our method vs. 22.44--30.12\,s for geometric NBV).

The comparison between \emph{Ours (w/o Stage II)} and \emph{Ours (Full)} shows that distance-aware refinement improves SR in four of five scenes and matches Stage I alone in Scene (a). Across 50 trials per configuration, the full system succeeds in 26 trials (52\%), compared with 20 trials (40\%) without Stage II. Both configurations record eight aborts and use the same mean number of Stage I views in each scene. Successes increase from two to four out of ten trials for the screwdriver and from three to five out of ten for the USB drive, with four USB aborts under each configuration.

Scene (c) remains the most challenging for our method, with 4/10 successes, 4/10 grasp failures, and 2/10 aborted trials. The screwdriver's thin geometry may limit feasible grasps and demand precise parallel-jaw execution, while the remaining aborted trials also indicate unresolved exploration failures.

\subsection{Ablation Studies}
To thoroughly evaluate the individual contributions of our proposed components and justify our architectural choices, we conduct extensive ablation studies in simulation. We analyze the design trade-offs in both the first-stage global NBV planner and the second-stage local view scorer.

Table~\ref{tab:ablation_stage1} compares the teacher VLM and distilled students. Directly using the VLM requires the fewest views (2.15) but incurs prohibitive latency (693 ms). A lightweight ResNet-18 reduces latency to 17 ms but increases required views (3.19). DINOv3-Base achieves a near-teacher view efficiency (2.26) with practical latency (28 ms), offering the best trade-off between semantic capability and real-time performance.

\begin{table}[tb]
\centering
\small
\caption{Ablation of Stage I Distilled Models. Evaluating the trade-off between search efficiency (\#Views) and inference latency (Avg  Frame Time).}
\label{tab:ablation_stage1}
\begin{tabular}{l @{\hspace{12pt}} c @{\hspace{12pt}} c}
\toprule
\textbf{Distilled Model} & \textbf{\#Views $\downarrow$} & \textbf{Avg Frame Time (ms) $\downarrow$} \\
\midrule
VLM Teacher (No Distill) & \textbf{2.15} & 693 \\
ResNet-18                & 3.19 & \textbf{17}   \\
ViT-Base                 & 2.60 & 23  \\
\textbf{DINOv3-Base (Ours)} & 2.26 & 28  \\
\bottomrule
\end{tabular}
\end{table}

Table~\ref{tab:ablation_stage2} evaluates feature design. Adding depth statistics and distance improves ASR from 47.92\% to 49.58\%. Incorporating object category prior further boosts performance (+9.72\% Seen, +7.29\% Unseen), indicating the importance of scale-aware distance modeling. Although adding affordance features slightly increases ASR (65.91\%), it raises latency above 210 ms, violating real-time constraints. Using DINOv3-Base without heavy affordance features achieves competitive performance (65.62\% / 58.42\%) under 48 ms latency, satisfying high-frequency refinement requirements.

\begin{table}[tb]
\centering
\small
\setlength{\tabcolsep}{3pt}
\caption{Ablation of Stage II Input Features (Baseline: ManipLLM). \textit{Img}: Image, \textit{Dep}: Depth, \textit{Dist}: Distance, \textit{Cat}: Category, \textit{Aff}: Affordance.}
\label{tab:ablation_stage2}
\begin{tabular}{l|ccc}
\toprule
\textbf{Scorer Configuration} & \textbf{Test Seen} & \textbf{Test Unseen} & \textbf{Avg Frame Time} \\
& \textbf{ASR (\%)} & \textbf{ASR (\%)} & \textbf{(ms)} \\
\midrule
ResNet-18 \textit{(Img)}                            & 47.92 & 44.09 & \textbf{18.82} \\
+ \textit{Dep} + \textit{Dist}                      & 49.58 & 46.77 & 26.10 \\
+ \textit{Dep} + \textit{Dist} + \textit{Cat}       & 59.30 & 54.06 & 27.97 \\
+ \textit{Dep} + \textit{Dist} + \textit{Cat} + \textit{Aff} & \textbf{65.91} & \textbf{59.31} & 210.68 \\
\midrule
\textbf{DINOv3-Base} & \multirow{2}{*}{65.62} & \multirow{2}{*}{58.42} & \multirow{2}{*}{47.82} \\
\textbf{+ \textit{Dep} + \textit{Dist} + \textit{Cat} (Ours)} & & & \\
\bottomrule
\end{tabular}
\end{table}

\begin{figure}[!ht]
    \centering
    \includegraphics[width=\linewidth]{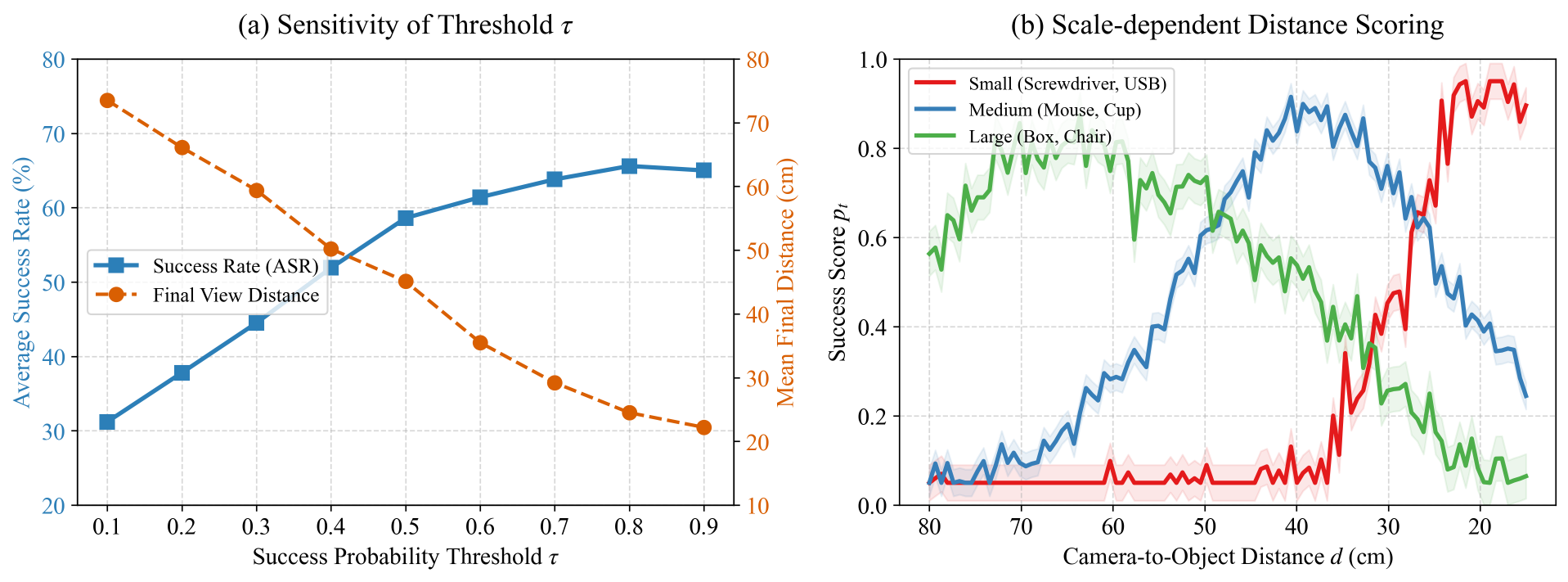}
    \caption{\textbf{Ablation on success threshold and object scale.} 
(a) Increasing the success threshold $\tau$ forces closer observation, improving ASR until saturation at 65.6\%. 
(b) Learned success score $p_t$ versus camera-to-object distance $d$ across object scales. Small objects require near-field observations, whereas large objects peak at intermediate distances and degrade under extreme proximity.}
    \label{fig:ablation_study_results}
\end{figure}

Fig.~\ref{fig:ablation_study_results}(a) shows that increasing $\tau$ reduces final distance (73.5 cm → 24.8 cm) and improves ASR until saturation at $\tau \approx 0.8$.

Fig.~\ref{fig:ablation_study_results}(b) illustrates scale-dependent behavior: small objects require near-field observations ($d < 30$ cm), whereas large objects achieve peak confidence at intermediate distances ($d \approx 62$ cm). This validates the necessity of explicit distance-aware modeling.

\section{Conclusion}

We presented DiNBV-Grasp, a distance-aware two-stage Next-Best-View framework that explicitly models the impact of camera--object distance on grasp success. Unlike conventional NBV methods that rely on a single global viewpoint, our approach extends active perception into the manipulator approach phase through continuous distance-aware view evaluation. By decoupling global semantic exploration from lightweight local refinement, the framework balances VLM-based reasoning capability with strict real-time constraints and can be integrated into existing object-centric grasping pipelines. Extensive simulation and real-world experiments demonstrate consistent improvements in grasp success across diverse object categories, with notable gains on small and partially occluded objects. Ablations show that distance cues complement RGB-D for view selection across object scales and sensing conditions. Future systems should jointly consider visibility, geometry, execution feasibility, and distance-dependent sensing quality.

\section*{Acknowledgements}
This work is supported by the National Natural Science Foundation of China (Grant No. 62502159, W2521174, 62476090, U23A20343, 62302167, 62222602), Fundamental Research Funds for the Central Universities, Natural Science Foundation of Shanghai (Grant No. 25ZR1402135),  the Shanghai Committee of Science and Technology (Grant No. 25511103300,
25511104302, 25511102700), ``Chen Guang'' project supported by Shanghai Municipal Education Commission and Shanghai Education Development Foundation (Grant No. 25CGA22), Natural Science Foundation of Chongqing (Grant No. CSTB2025NSCQ-GPX0445), Young Elite Scientists Sponsorship Program by CAST (Grant No. YESS20240780), Open Project Program of the State Key Laboratory of CAD\&CG (Grant No. A2501), Zhejiang University, Open Research Fund of Key Laboratory of Advanced Theory and Application in Statistics and Data Science-MOE, ECNU.

\clearpage
%
%
\bibliographystyle{splncs04}
\bibliography{main}
\end{document}


\title{Supplementary Material for DiNBV-Grasp} 

\titlerunning{DiNBV-Grasp}

\author{Zilong Xie\inst{1}\orcidlink{0009-0001-4436-5322} \and
Jingyu Gong\inst{1,2,3}\orcidlink{0000-0002-4536-0953} \and
Xin Tan\inst{1,2}\orcidlink{0000-0001-9346-1196} \and
Zhizhong Zhang\inst{1,3}\orcidlink{0000-0001-6905-4478} \and
Yanyun Qu\inst{4}\orcidlink{0000-0002-8926-4162} \and
Lizhuang Ma\inst{1}\orcidlink{0000-0003-1653-4341} \and
Yuan Xie\inst{1,2}\orcidlink{0000-0001-6945-7437}}

\authorrunning{Z.~Xie et al.}

\institute{School of Computer Science and Technology, East China Normal University, China \and
Chongqing Key Laboratory of Precision Optics, Chongqing Institute of East China Normal University, China\and
Shanghai Key Laboratory of Computer Software Evaluating and Testing, China\and
School of Informatics, Xiamen University, China\\
\email{zilong\_6037@stu.ecnu.edu.cn} \email{yyqu@xmu.edu.cn}\\
\email{\{jygong, xtan, zzzhang, lzma, yxie\}@cs.ecnu.edu.cn}}

\maketitle

\noindent This supplementary material provides implementation and evaluation details that complement the main paper. It describes the two-stage execution logic, the visibility-adaptive NBV utility, teacher-label generation for semantic distillation, distance-aware feature construction, sensor-related design choices, and additional analysis protocols.

\section*{A. Additional Pipeline Details}

DiNBV-Grasp separates active perception into two stages with different computational roles. Stage I performs global semantic exploration from a discrete set of collision-free candidate views sampled around a fixed tabletop search region, following the general active-view-planning paradigm used in recent NBV-based grasping systems~\cite{lauri2020multi,ma2024active,wang2025graspview,shi2025viso}. At each step, the current scene representation is updated from the latest RGB-D observation, candidate views are rendered, and the next-best-view is selected using geometric, semantic, and execution feasibility scores. In our experiments, Stage I stops when exploration reaches $N_{\max}=5$ steps after $I_0$ or target visibility $V_t$ reaches 0.75. Stage II begins only if the selected observation yields a feasible initial grasp hypothesis $\pi_0$.

Stage II runs during the manipulator approach. The robot scores intermediate RGB-D observations and keeps the highest-scoring observation seen so far. Sampling stops when $\max_t p_t \ge \tau$ or the camera reaches the minimum observation distance $d_{\min}$. The final grasp is synthesized from the highest-scoring observation acquired by the stopping step. The complete pipeline is summarized in Table~\ref{tab:overall_procedure}.

\begin{table}[t]
\centering
\caption{Overall procedure of the proposed grasping framework.}
\label{tab:overall_procedure}
\renewcommand{\arraystretch}{1.2}
\begin{tabular}{p{0.98\linewidth}}
\toprule
\textbf{Overall Procedure} \\
\midrule
\textbf{Input:} target instruction $T$, initial RGB-D observation $I_0$, candidate view set $\mathcal{V}$.\\
\textbf{Output:} final grasp pose $\pi^\star$ or safe abort status.\\[2mm]

Initialize $\mathcal{S}$ from $I_0$, evaluate $V_0$, and set $t=0$.\\

\textbf{while} $t<N_{\max}$ \textbf{and} $V_t<0.75$ \textbf{do}\\
\hspace*{1em}Render candidate observations $\{\tilde I_v\}_{v\in\mathcal{V}}$ from $\mathcal{S}$.\\
\hspace*{1em}Remove views that violate IK, collision, or safety constraints.\\
\hspace*{1em}Compute $S_{\text{geo}}(v)$, $S_{\text{sem}}(v)$, and $S_{\text{exec}}(v)$ for each feasible view.\\
\hspace*{1em}Select
$v_t^\star=\arg\max_{v\in\mathcal{V}}S_{\text{NBV}}(v,t)$
and move the robot to $v_t^\star$.\\
\hspace*{1em}Acquire a new RGB-D observation and update $\mathcal{S}$ and $V_t$.\\
\hspace*{1em}$t\leftarrow t+1$.\\
\textbf{end while}\\

Generate an initial grasp hypothesis $\pi_0$ from the selected global observation, if available.\\
\textbf{if} no valid observation or feasible $\pi_0$ is available \textbf{then}\\
\hspace*{1em}Return safe abort status.\\
\textbf{end if}\\

\textbf{while} approaching the target and $d_t>d_{\min}$ \textbf{do}\\
\hspace*{1em}Acquire intermediate observation $(I_t,D_t)$ and distance $d_t$.\\
\hspace*{1em}Compute the local score $p_t$ with the distance-aware scorer.\\
\hspace*{1em}Keep the observation with the highest $p_t$.\\
\hspace*{1em}\textbf{if} $\max_t p_t\ge\tau$ \textbf{then break}.\\
\textbf{end while}\\

Synthesize and execute the final grasp $\pi^\star$ from the selected observation.\\
\bottomrule
\end{tabular}
\end{table}

\section*{B. Visibility-Aware Dynamic Weighting}

At Stage I step $t$, each feasible candidate view $v\in\mathcal{V}$ is scored by a weighted utility:
\begin{equation}
S_{\text{NBV}}(v,t)
=
\alpha_t S_{\text{geo}}(v)
+
\beta_t S_{\text{sem}}(v)
+
\gamma_t S_{\text{exec}}(v),
\end{equation}
where all score terms are normalized to $[0,1]$ over the candidate set at the current step. The weights depend on the current target visibility $V_t$, so the planner can emphasize exploration when the target is poorly observed and gradually shift toward semantic and execution-aware view selection as the target becomes visible.

\noindent\textbf{Target visibility.}
Before exploration, independent unoccluded RGB-D calibration samples provide a reference mask area $A_{\text{ref}}(T)$ at depth $z_{\text{ref}}$ for the target category. If SAM3~\cite{carion2025sam} detects no target mask in the current observation, we set $V_t=0$. Otherwise, let $|M_t|$ be the mask area in pixels and $z_t$ the target depth measured within it. For fixed camera intrinsics, we use
\begin{equation}
V_t
=
\mathrm{clip}
\left(
\frac{|M_t|}{A_{\text{ref}}(T)}
\left(\frac{z_t}{z_{\text{ref}}}\right)^2,\,
0,\,
1
\right),
\end{equation}
a depth-corrected visibility proxy that remains sensitive to object orientation and within-category size variation.

\noindent\textbf{Geometric score.}
Within the fixed tabletop search region, accumulated depth observations mark traversed voxels as free and measured surfaces as occupied. Unobserved voxels remain unknown. From each candidate $v$, we cast rays within the camera's field of view and sensing range, stopping each ray at the first occupied or unknown voxel. Let $F_t(v)$ contain the distinct unknown voxels reached this way. The geometric score is
\begin{equation}
S_{\text{geo}}(v)
=
\mathrm{Norm}_{v\in\mathcal{V}}
\bigl(|F_t(v)|\bigr).
\end{equation}
If all counts are equal, the geometric term is set to zero for all candidates.

\noindent\textbf{Semantic score.}
Until the target is first detected in an acquired RGB-D observation, all candidates receive $S_{\text{sem}}(v)=0$. Thereafter, the distilled student scores the rendered views~\cite{wang2025graspview}:
\begin{equation}
S_{\text{sem}}(v)
=
f_{\phi}(\tilde I_v,T),
\end{equation}
where $\tilde I_v$ is the rendered candidate image and $T$ is the target instruction. The teacher supervision for $f_{\phi}$ evaluates target visibility, semantic relevance, and graspability, as described in Sec.~C.

\noindent\textbf{Execution score.}
The execution term filters candidate views that are visually useful but physically difficult to realize:
\begin{equation}
S_{\text{exec}}(v)
=
\mathbf{1}_{\text{IK}}(v)\,
\mathbf{1}_{\text{col}}(v)\,
\mathbf{1}_{\text{safe}}(v)\,
\exp(-\kappa c_{\text{motion}}(v)),
\end{equation}
where $\mathbf{1}_{\text{IK}}$, $\mathbf{1}_{\text{col}}$, and $\mathbf{1}_{\text{safe}}$ indicate inverse-kinematics feasibility, collision-free motion, and camera/manipulator safety constraints. $c_{\text{motion}}(v)$ is the normalized motion cost from the current pose to $v$, and $\kappa$ controls how strongly long motions are penalized.

\noindent\textbf{Weight schedule.}
We use a piecewise schedule:
\begin{equation}
(\alpha_t,\beta_t,\gamma_t)
=
\begin{cases}
(\alpha_{\text{hi}},\beta_{\text{lo}},\gamma_{\text{lo}}), & V_t < 0.2,\\
(\alpha_{\text{mid}},\beta_{\text{mid}},\gamma_{\text{mid}}), & 0.2 \le V_t < 0.75.
\end{cases}
\end{equation}
The first regime prioritizes geometric discovery under severe occlusion. The second regime balances geometric coverage, semantic target relevance, and executable motion. Once $V_t\ge0.75$, Stage I stops. Local refinement then requires a feasible initial grasp hypothesis.

\noindent\textbf{View selection.}
At each exploration step, infeasible candidates are removed first, the remaining score terms are normalized across candidates, and the robot executes
\begin{equation}
v_t^\star
=
\arg\max_{v\in\mathcal{V}} S_{\text{NBV}}(v,t).
\end{equation}
After executing $v_t^\star$, the new RGB-D observation updates the scene representation and the process repeats until the Stage I stopping condition is met.

\section*{C. Teacher VLM Scoring and Prompt Template}

The teacher model is Qwen3-VL-32B-Instruct~\cite{bai2025qwen3}. For each rendered candidate view, the teacher receives the target instruction and the rendered RGB image, then outputs a scalar supervision score in $[0,1]$. The score reflects whether the candidate view is useful for revealing and grasping the target object.

\noindent\textbf{Prompt template.}
\begin{quote}
You are evaluating a candidate camera view for robotic grasping. The target instruction is: \texttt{\{target instruction\}}. Given the image, score the view from 0 to 1. Consider: (1) whether the target object is visible, (2) whether the target is semantically identifiable and not confused with nearby objects, and (3) whether the visible geometry appears useful for grasp planning. Return only a single numeric score between 0 and 1.
\end{quote}

The distilled student is trained to approximate this scalar teacher score for rendered candidate views. At runtime, the student replaces repeated teacher calls while preserving the semantic ranking behavior needed for NBV selection.

\section*{D. Distance-Aware Scorer and Depth Processing}

The Stage II scorer receives RGB features, low-dimensional depth statistics, normalized camera-object distance, and a CLIP text embedding of the object category~\cite{radford2021learning}. The model predicts the probability that a grasp generated from the current observation will succeed.

The distance term complements the compact depth statistics by retaining information about the camera--object distance. RGB-D sensors such as the RealSense D435i also exhibit distance-dependent noise: near-field measurements can contain missing-depth artifacts, while far-field observations lose fine geometric detail. Since the useful observation distance varies with object scale, the explicit distance feature helps the scorer learn a scale-aware perception-quality metric.

We compute the 10-dimensional depth statistics from numeric depth measurements. Colorized depth maps shown in figures are used only for visualization. Independent multi-distance grasp executions provide observation-level binary labels. We fine-tune on a few real-world images we collected.

\section*{E. Execution Status and Safety Handling}

DiNBV-Grasp maintains an execution status during global exploration and local approach. If Stage I ends without a valid observation or a feasible initial grasp hypothesis, the trial is aborted and the robot returns to a safe state, consistent with closed-loop grasp planning systems that explicitly handle failed observation or grasp hypotheses~\cite{breyer2022closed,Contactgraspnet2021}.

The current action space is limited to viewpoint adjustment and prehensile grasping. Non-prehensile rearrangement actions, such as pushing or sweeping surrounding objects, are not included in the execution policy.

\section*{F. Evaluation Splits and Sensor Considerations}

The simulation evaluation uses two splits. Test Seen contains novel instances from categories observed during training, while Test Unseen contains novel categories excluded from discriminator training. This split design evaluates both instance-level robustness and category-level generalization.

In real-world experiments, depth sensing is affected by missing values, reflective surfaces, near-field noise, and camera-object distance. We use $d_{\min}=15$ cm as the minimum observation distance during approach in this setup. The scorer also uses depth statistics rather than colorized depth screenshots, which reduces sensitivity to visualization artifacts and better reflects the numeric sensor signal.

\section*{G. Sensitivity Analysis Details}

The success threshold $\tau$ controls when Stage II stops approaching and commits to the best observation. Increasing $\tau$ generally encourages closer inspection and improves success until saturation around $\tau=0.8$, after which sensing artifacts and kinematic constraints limit further gains.

We use $d_{\min}=15$ cm as the stopping distance for Stage II observation sampling in our setup.

The ranking weights $\lambda_1$ and $\lambda_2$ balance sample-level calibration and distance ordering. We use $\lambda_1=1.0$ and $\lambda_2=0.5$ with margin $m=0.2$ to encourage observations associated with successful outcomes to score higher than those associated with failed outcomes from the same scene.

\section*{H. Additional SAPIEN Qualitative Results}

\begin{figure}[t]
\centering
\includegraphics[width=\linewidth]{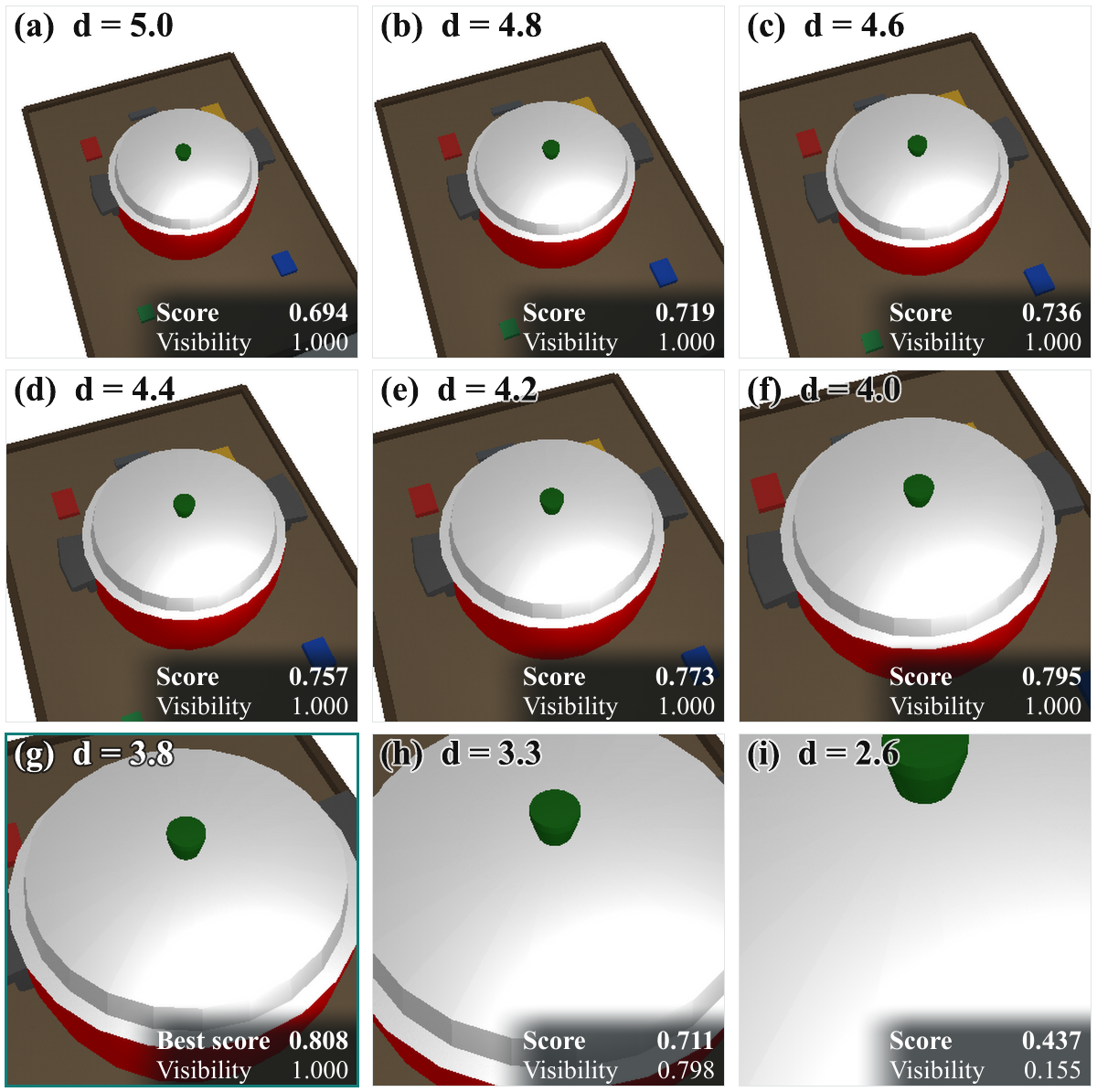}
\caption{
Additional SAPIEN qualitative result for the KitchenPot target. Candidate observations are ordered from far to near, and the highlighted view is selected by the Stage II distance-aware scorer. Here, $d$ is the camera orbit radius in decimeters (dm).
}
\label{fig:raw_distance_sequence_111}
\end{figure}

\begin{figure}[t]
\centering
\includegraphics[width=\linewidth]{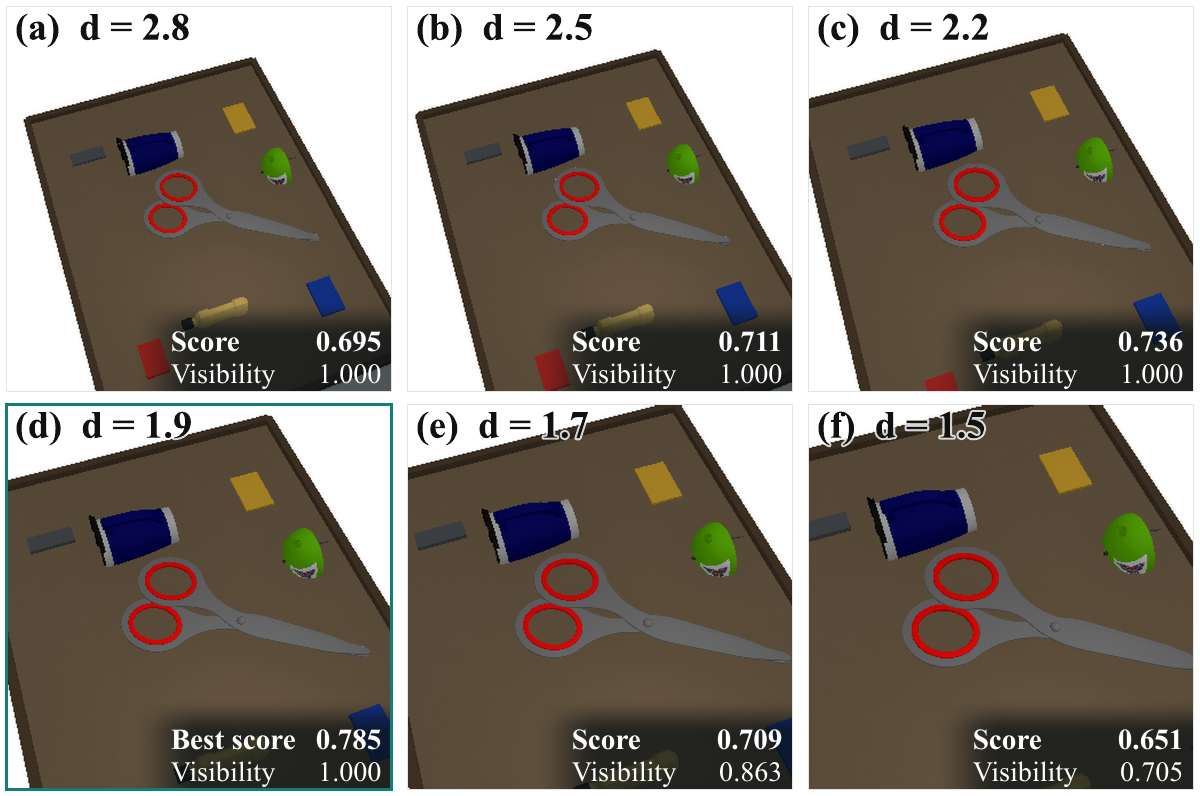}
\caption{
Additional SAPIEN qualitative result for the Scissors target. Six candidate observations are ordered from far to near, and the highlighted view reflects an intermediate orbit radius that preserves both object geometry and useful scene context. The distance labels $d$ are in decimeters (dm).
}
\label{fig:raw_distance_sequence_222}
\end{figure}

Figure~\ref{fig:raw_distance_sequence_111} and Figure~\ref{fig:raw_distance_sequence_222} provide additional qualitative examples of the Stage II distance-aware refinement process in the SAPIEN simulator~\cite{xiang2020sapien}. The figures show representative RGB observations sampled from a dense sweep of the camera orbit radius and arranged from far to near. The highlighted observation denotes the view selected by the distance-aware scorer, which predicts the grasp success probability from the current RGB-D observation, depth statistics, object-category embedding, and normalized camera-object distance.

These examples illustrate that the most useful observation is not necessarily the closest one. Farther views preserve more scene context but may lose fine object geometry, while overly close views can suffer from target cropping, reduced contextual cues, or unreliable local depth structure. The selected intermediate observations provide a better trade-off between geometric detail and observation stability. This qualitative behavior is consistent with the main paper: DiNBV-Grasp benefits from explicitly modeling camera-object distance during the approach stage, enabling the system to select an informative observation for final grasp synthesis rather than simply committing to the nearest available view.

\clearpage
\bibliographystyle{splncs04}
\bibliography{main}